\documentclass[a4paper,twoside]{article}

\usepackage{epsfig}
\usepackage{subcaption}
\usepackage{calc}
\usepackage{amsmath,amssymb,amsfonts}
\usepackage{amstext}
\usepackage{amsthm}
\usepackage{multicol}
\usepackage{pslatex}
\usepackage{apalike}
\usepackage{algorithm2e}
\usepackage[bottom]{footmisc}
\usepackage{bm}
\usepackage{enumitem}
\usepackage{algorithmic}
\usepackage{graphicx}
\usepackage{textcomp}
\usepackage{xcolor}
\usepackage{float}
\usepackage{latexsym}
\usepackage[export]{adjustbox}
\usepackage{makecell} 
\usepackage{url} 
\usepackage{SCITEPRESS}

\begin{document}

\title{A Novel Metric for Measuring Data Quality in Classification Applications (extended version)}

\author{\authorname{Jouseau Roxane, Salva Sébastien and Samir Chafik}
\affiliation{Université Clermont-Auvergne, CNRS, Mines de Saint-Etienne, Clermont-Auvergne-INP, LIMOS, Clermont-Ferrand, France}
\email{roxane.jouseau@doctorant.uca.fr, sebastien.salva@uca.fr, chafik.samir@uca.fr}
}


\abstract{Data quality is a key element for building and optimizing good learning models. Despite many attempts to characterize data quality, there is still a need for rigorous formalization and an efficient measure of the quality from available observations. Indeed, without a clear understanding of the training and testing processes, it is hard to evaluate the intrinsic performance of a model. Besides, tools allowing to measure data quality specific to machine learning are still lacking. In this paper, we introduce and explain a novel metric to measure data quality. This metric is based on the correlated evolution between the classification performance and the deterioration of data. The proposed method has the major advantage of being model-independent. Furthermore, we provide an interpretation of each criterion and examples of assessment levels. We confirm the utility of the proposed metric with intensive numerical experiments and detail some illustrative cases with controlled and interpretable qualities.}

\onecolumn \maketitle \normalsize \setcounter{footnote}{0} \vfill

\section{Introduction}
\label{introduction}
During the last decades, data availability has played a crucial role in the development and sophistication of artificial intelligence in general and machine learning models in particular. Yet, few works have been dedicated to deeply investigating the assessment on the apparent accuracy and consistency of data~\cite{DQFramework,methodologiesDataQ,DQbook,dataQAssesment}. Moreover, the term data quality has been restricted to studying the impact of some standard criteria on the task at hand and user's expectations. Standard criteria have been evaluated very often separately or with a non-rigorous combination of scores. For example, several methods use the term data quality for specific purposes, such as accuracy, completeness, or timeliness, for specific applications and contexts~\cite{DQMLBD}. Unfortunately, this leads to the first limitation related to data quality: The lack of an appropriate definition and, subsequently, an accurate measure.

Classical data evaluation is often related to a given context: External metadata, rules, trusted references, etc. Establishing or extracting these external elements is usually a long process, expert-dependent and error-prone~\cite{DQmeasurement}. Instead of developing new methods, some previous papers cite commercial products used to measure data quality. A common conclusion states that only a few tools are available. We believe that this is a consequence of the limitations detailed above. Nevertheless, we have tried to test some of them for the same context, but they were either out of the scope of our study, or we were unable to set up a usable configuration.

In this work, we solve the previous limitations by introducing an original data quality metric. We focus on the use of data in artificial intelligence applications and, more specifically, for learning models in numerical classification. The proposed metric has the significant advantage of providing a consistent means to evaluate the quality of different types of data with different numbers of classes, various domains, and from low to high dimensionality, etc. All the steps were constructed carefully to make the metric interpretable, easy to use, and model-independent. To illustrate how the metric can successfully capture a deterioration, we have shown that the impact makes the classification performance decrease non-linearly with different rates. This is against a non-justified prior when, in some previous works, they have assumed a linear behavior.

\subsection{Related Work}
\label{soa}

Only a few tools are available to directly evaluate the quality of a dataset with a metric, as most tools only offer indicators to monitor and help with data profiling. In the recent survey \cite{DQmeasurement} the authors evaluated $11$ tools: Apache Griffin \cite{ApacheGriffin}, Ataccama ONE \cite{Ataccama}, DataCleaner \cite{datacleaner}, Datamartist \cite{datamartist}, Experian Pandora \cite{experian}, InformaticaDQ \cite{informatica}, InfoZoom \& IZDQ \cite{infozoomIZDQ}, MobyDQ \cite{mobyDQ}, OpenRefine \cite{openrefine} \& MetricDoc \cite{metricdoc}, SAS Data Quality \cite{sasDQ}, and Talend Open Studio \cite{talendOSDQ}. These tools were classified into $5$ categories: accuracy, completeness, consistency, timeliness, and others. Interestingly, most of these tools allow the evaluation of one or two criteria only. They mainly focus on easing the definition of data indicators and assisting data profiling. Ultimately, only $4$ tools were distinguished: Apache Griffin, InformaticaDQ, MobyDQ, and MetricDoc. 

While many of these tools focus on the attribute level, they are not generalized for higher levels of aggregation. They also considered data rules, which are not considered here because they require expert knowledge and are often unavailable. Additionally, Apache Griffin and MobyDQ require a reference dataset. This makes them less practical as ground truth reference datasets are not always available. InformaticaDQ focuses on textual data, such as elements of postal addresses, email addresses, etc., and cannot be applied to numerical data. MetricDoc offers two different time interval metrics for time-series data, a redundancy metric on the table level and metrics for validity and plausibility, both defined at the attribute level only. Finally, we tried to investigate the Data Quality for AI API proposed by IBM \cite{ibmDQ}. However, despite taking steps to access the free trial version on the website, we could not secure working access to the API, which is a problem that the authors of \cite{DQmeasurement} also seem to have faced, as mentioned in their paper.

\subsection{Contributions}
In order to build a metric that is independent of learning models, we formulate the problem such that no reference or expertise is needed. Instead, our metric is based upon two main terms: The former evaluates classification performance across a wide range of models, and the latter assesses variations of performance when a low amount of errors is injected into datasets. We show that high variations are consequences of quality issues. All terms are then empirically combined to form a unique evaluation, denoted  $q_a$. Furthermore, we show how to interpret our metric scores and express the notions of good, medium, and bad qualities. Then, we evaluate the metric $q_a$ on $110$ datasets of known quality and show that $q_a$ is able to measure the quality correctly. We also discuss the information that can captured by $q_a$, leading to an easy connection with a given result. 

To the best of our knowledge, this is the first method that proposes a clear and rigorous formulation for measuring data quality. We summarize the main contributions of this work as follows: We propose a novel rigorous metric to evaluate data quality for numerical classifications. Our method can be adapted and extended to regression. The proposed metric is model-independent. This proposed metric does not require external elements or expert supervision. We use a constructive approach, step by step, to formulate the metric that could be generalized to other contexts. 

The rest of the paper is organized as follows: Section \ref{qaintro} proposes a definition for the data quality metric and some thresholds to ease interpretation. Then,  the metric is evaluated on $155$ datasets of varied levels of quality in different contexts. Possible threats to validity are also discussed in Section \ref{EmpEval}. Finally, we conclude the paper in Section \ref{conclusion}.

\section{Measuring Data Quality}
\label{qaintro}
In \cite{IDEAL}, we investigated the relevance of repairing datasets according to the impact of the amount of errors included in datasets on the classification performance. This study allowed the observation of some distinct characteristics related to data quality. In particular, we observed the two following properties:
\begin{enumerate}[noitemsep,topsep=0pt]
    \item Model accuracy decreases along with data quality when errors are injected into data. This decrease varies across classification models and numbers of classes;
    \item The decrease in accuracy is nonlinear. It is low when data is of good quality or when data is extremely deteriorated. However, the decrease in accuracy is significantly higher between these two states.
\end{enumerate}

\begin{figure*}[!h]
    \hspace{-0.5cm}
    \centering
    \begin{subfigure}{0.35\textwidth}
        \includegraphics[width=\textwidth]{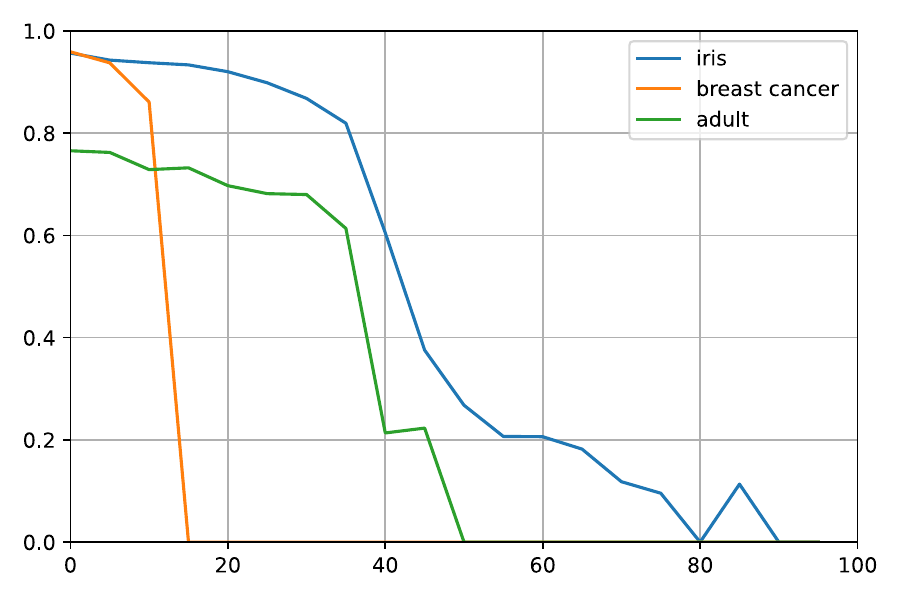}
        \vspace*{-0.6cm}
        \caption{}
        \label{accm}
    \end{subfigure}
    \hspace{-0.35cm}
    \begin{subfigure}{0.35\textwidth}
        \includegraphics[width=\textwidth]{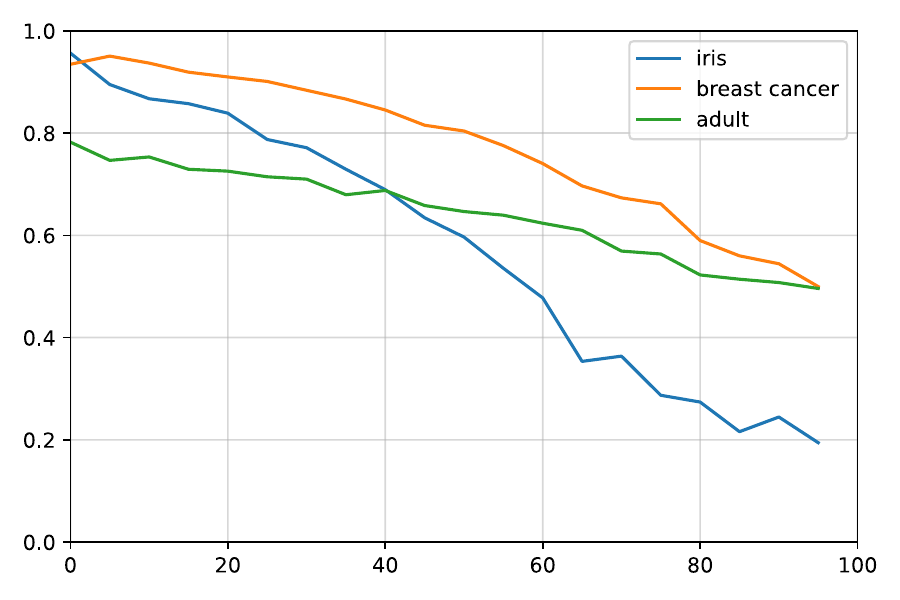}
        \vspace*{-0.6cm}
        \caption{}
        \label{acco}
    \end{subfigure}
    \hspace{-0.35cm}
    \begin{subfigure}{0.35\textwidth}
        \includegraphics[width=\textwidth]{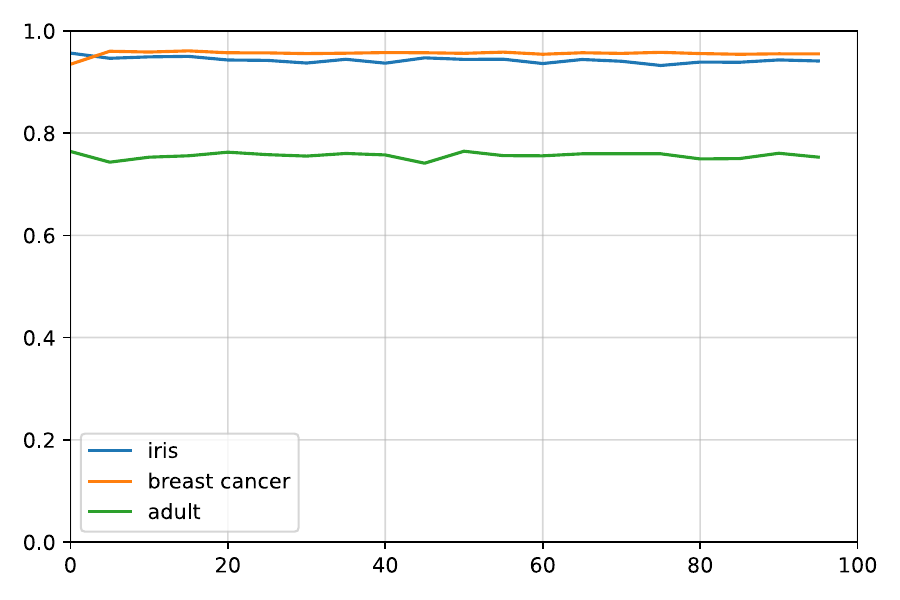}
        \vspace*{-0.6cm}
        \caption{}
        \label{accf}
    \end{subfigure}
    \hspace{-0.5cm}
    \caption{Evolution of mean accuracy when errors are injected in training. Missing values (a), outliers (b), and fuzzing (c).}
    \label{errorinject}
    \vspace{-0.2cm}
\end{figure*}

We illustrate these observations on three datasets (Iris, Breast cancer, Adult). We chose 12 standard classification models available in \cite{scikit-learn}: Logistic regression, K-Nearest Neighbors, Decision tree, Random forest, Ada boost, Naive Bayes, XGboost, Support vector classification, Gaussian process, Multi-layer perceptron, Stochastic gradient descent, and Gradient boosting. We also illustrate these observations for the error types missing values, outliers, and fuzzing, a.k.a. partial duplicates. We chose these error types because we observed that they have different impacts on model performance: Outliers and missing values have the most impact on accuracies and f1 scores; fuzzing tends to have less impact and offers the benefit of simulating data generation. In Figure \ref{errorinject}, we present mean accuracies computed over 30 iterations of injecting controlled percentages of errors, randomly generated with a uniform distribution, in training data. We inject up to 95\% of errors with a 5\% increment. The 12 classification models are then trained on these deteriorated datasets with a random split using 80\% for training and 20\% for testing. Figures \ref{accm}, \ref{acco}, and \ref{accf} present the mean accuracies when missing values, outliers, and fuzzing are respectively injected in training data.

In Figures \ref{accm} and \ref{acco}, as per our first observation, accuracies are high when none or a low percentage of errors are present in the data. Unsurprisingly, these accuracies decrease as data deterioration increases. But, as we stated in our second observation, the decrease in accuracy is non-linear. This is especially visible in Figure \ref{accm}. We can also see in Figures \ref{accm} and \ref{acco} that the mean accuracy can drop significantly with only a 5\% increment of errors. For instance, in Figure \ref{accm}, for the dataset Iris, the mean accuracy stays over 0.8 up to the injection of 35\% of missing values into training data. However, at 40\% of missing values, the mean accuracy drops to 0.6, approximately. The data fuzzing observed in Figure \ref{accf} is a particular case with no loss of information, which is why accuracies stay quite steady. 

The analysis and formalization of these observations allow us to propose the first data quality measurement using the notion of data deterioration. This metric, denoted $q_a$, is composed of two parts, denoted $q_{a,1}$ (Eq. \ref{basepartacc}) and $q_{a,2}$ (Eq. \ref{varpartacc}), which respectively encode these characteristics:
 
\begin{itemize}[noitemsep,topsep=0pt]
    \item $q_{a,1}$: the accuracies across a set of classification models (Observation 1). $q_{a,1}$ also accounts for the number of classes in datasets in order to allow comparisons of data quality levels between datasets with different numbers of classes;
    
    \item $q_{a,2}$: variations of accuracies when a low percentage of errors are injected in training sets (Observation 2). It aims to capture abnormally high accuracy variations over small dataset perturbations.
\end{itemize}
Next, we formalize these two observations in a rigorous way.

\subsection{Definition of $\mathrm{\mathbf{q_{a,1}}}$}
\label{qa1}
We use the following notations in the remainder of the paper: The set of models is denoted $M$, the set of error types is denoted $E$, and $D$ is the dataset under evaluation. $A(m,D)$ stands for the accuracy of the model $m \in M$ on D. We define $A_M(D)$ as :
\abovedisplayskip=3pt
\belowdisplayskip=3pt
\begin{equation}
    A_M(D) := \frac{1}{card(M)}\sum_{m \in M} A(m,D)
\end{equation}
Observing the mean accuracy $A_M(D)$ alone is not sufficient to express data quality. Moreover, given the number of classes $c$ in $D$, we want an accuracy that is better than a random choice:
\abovedisplayskip=3pt
\belowdisplayskip=3pt
\begin{equation}
    1 \geq A_M(D) > \frac{1}{c} \ , \ c>1
\end{equation}

When the accuracy of a model is lower than $\frac{1}{c}$, we consider the quality to be the lowest. The function $\delta_1$ capture this statement:
\abovedisplayskip=3pt
\belowdisplayskip=3pt
\begin{eqnarray}
    \delta_1(A_M(D)) :=
    \begin{cases}
    1 & if\  A_M(D) > \frac{1}{c}\\
    0 &\ otherwise
    \end{cases}
    \label{delta1}
\end{eqnarray}

Finally, we define $q_{a,1}$ as:
\abovedisplayskip=3pt
\belowdisplayskip=3pt
\begin{equation}
    q_{a,1}(D) := 1 - \frac{c A_M(D) - 1}{c - 1}\delta_1(A_M(D)) \quad
    \label{basepartacc}
\end{equation}
We have $0 \leq q_{a,1}(D) \leq 1$, with $q_{a,1}(D)=0$ as the best quality.

\subsection{Definition of $\mathrm{\mathbf{q_{a,2}}}$}
\label{qa2}

The main idea encoded by $q_{a,2}$ is that the mean accuracies of models trained on good or bad-quality datasets are not sensitive to a small data deterioration. This is not the case for datasets of medium quality. We verify this hypothesis by computing variations of accuracy with the set of classification models $M$ when a percentage $p$ of an error type $e \in E$ is injected in $D$. We denote $D_{e,p}$ the resulting dataset.

For $\Delta A_{M,e}(D)$ to expresses the variation of accuracy for a specific error $e$, we introduce:
\abovedisplayskip=3pt
\belowdisplayskip=3pt
\begin{equation}
    \Delta A_{M,e}(D) := \frac{1}{card(M)}\sum\limits_{m \in M}|A(m,D) - A(m,D_{e,p})|
    \label{deltaAe}
\end{equation}

To avoid any bias in our metric, we assume that errors are injected randomly, with a uniform distribution in training data. In our experiments, we noticed in Figure \ref{accm} and \ref{acco} that a small percentage of errors $p=5\%$ is sufficient to capture accuracy variations. According to our experiments, a higher value of $p$ is possible but may lead to less precise measurements of the performance loss. The goal of this parameter is to simulate small perturbations in data that can happen in real-life situations. Although small variations of accuracies are expected, we are only interested in abnormal variations. To exclude these minor variations, we define $\delta_2$ as:
\abovedisplayskip=3pt
\belowdisplayskip=3pt
\begin{eqnarray}
    \delta_2(\Delta A_{M,e}(D)) :=
    \begin{cases}
        1 &\ if\ \Delta A_{M,e}(D) > p\\
        0 &\ otherwise
    \end{cases}
    \label{delta2}
\end{eqnarray}
$q_{a,2}$ is then defined as:
\abovedisplayskip=3pt
\belowdisplayskip=3pt
\begin{equation}
q_{a,2}(D) := min(\frac{10}{card(E)}\sum\limits_{e \in E} \Delta A_{M,e}(D)\delta_2(\Delta A_{M,e}(D)), 1)
    \label{varpartacc}
\end{equation}

We chose to add a factor of $10$ in $q_{a,2}(D)$ because we consider that an accuracy variation of $10\%$ or more when we inject a small amount of error in data, indicates bad data quality. However, this factor does not keep the result bounded by 1, so we use the minimum function to define $q_{a,2}$. It is worth noting that this parameter may be easily changed to meet user preferences.

\subsection{The Impact of Untrusted Test Data}
\label{badtestdata}
\begin{figure*}[!h]
    \hspace{-0.5cm}
    \centering
    \begin{subfigure}{0.35\textwidth}
        \includegraphics[width=\textwidth]{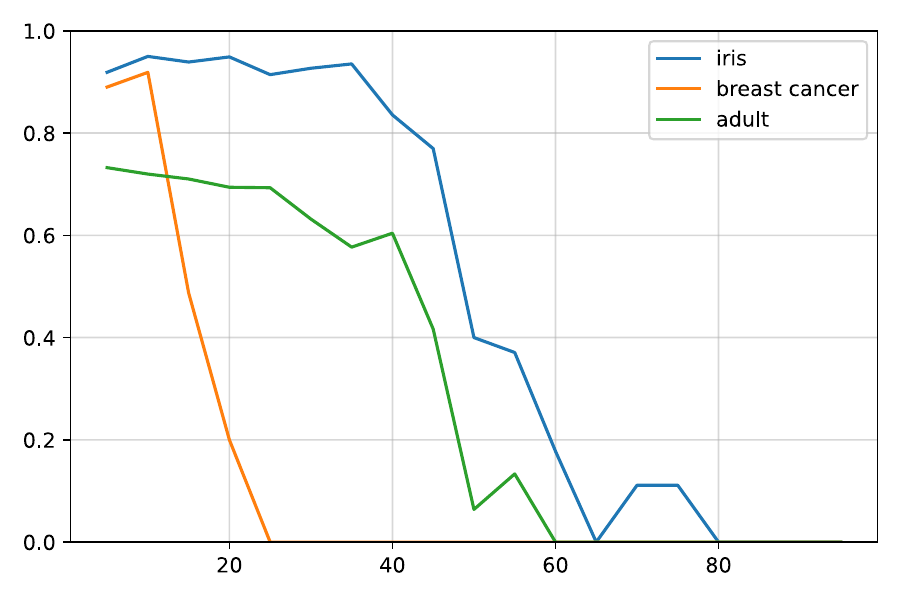}
        \vspace*{-0.6cm}
        \caption{}
        \label{badm}
    \end{subfigure}
    \hspace{-0.35cm}
    \begin{subfigure}{0.35\textwidth}
        \includegraphics[width=\textwidth]{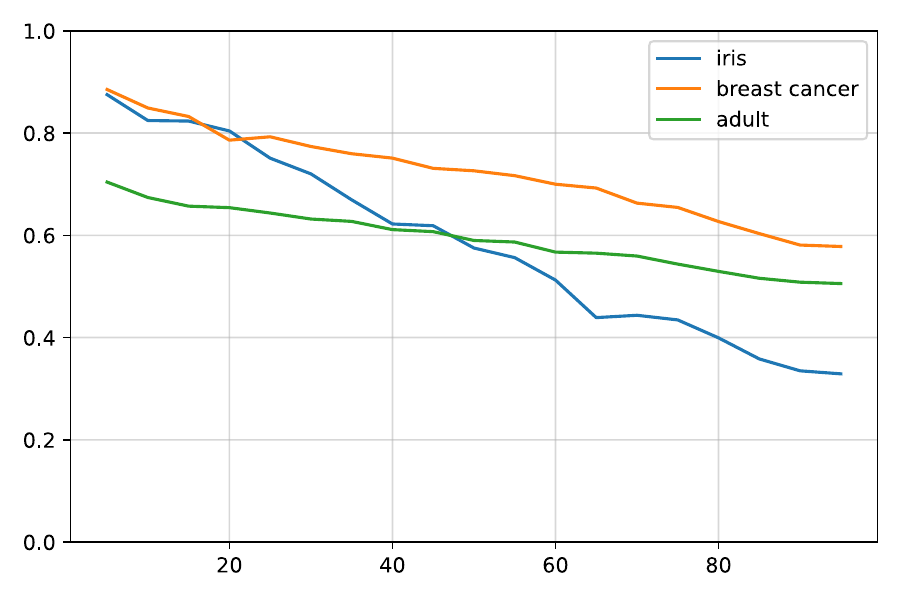}
        \vspace*{-0.6cm}
        \caption{}
        \label{bado}
    \end{subfigure}
    \hspace{-0.35cm}
    \begin{subfigure}{0.35\textwidth}
        \includegraphics[width=\textwidth]{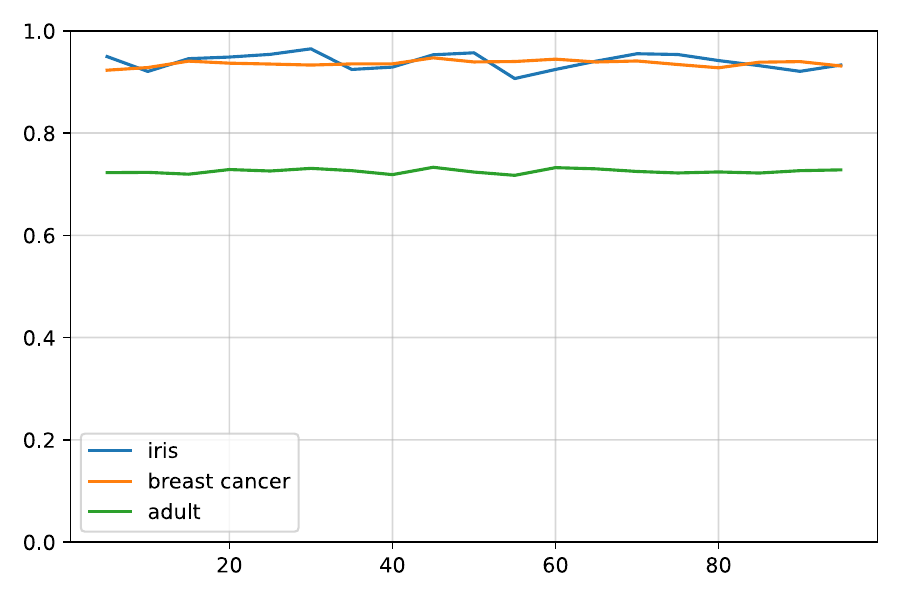}
        \vspace*{-0.6cm}
        \caption{}
        \label{badf}
    \end{subfigure}
    \hspace{-0.5cm}
    \caption{Evolution of mean accuracy when errors are injected in data: missing values (a), outliers (b), and fuzzing (c).}
    \label{badTest}
    \vspace{-0.2cm}
\end{figure*}

The definitions of $q_{a,1}$ and $q_{a,2}$ are based on the computation of accuracies and thus rely on the existence of trusted test data. However, in real-life situations, test data is often a sample of the dataset and, therefore, of the same quality. To ensure whether $q_{a,1}$ and $q_{a,2}$ are valid in this scenario, we injected controlled percentages of errors (missing values, outliers, and fuzzing) from 0\% to 95\% with a 5\% increment in the entire three previous datasets. We then extracted test and training sets with random samplings. To lower the chances of sampling an unreliable test set, we repeated the random sampling and training 30 times. Then, we trained the 12 classification models listed in Section \ref{qaintro}. In Figure \ref{badTest}, we present, for each error type, the mean accuracies of the 12 classification models over the 30 random samplings of training and test sets obtained from Iris, Breast cancer, and Adult. If we compare the figures \ref{accm} and \ref{badm}, we observe that even though individual values of mean accuracies are different, they stay in the same range. And most importantly, the overall decreases in mean accuracies reflect the same tendencies. This is also true when we compare the figures \ref{acco}, \ref{bado} and the figures \ref{accf}, \ref{badf}. We, therefore, consider that using a mean of $q_{a,1}$ and $q_{a,2}$ over 30 resamplings when trusted test data is not available is appropriate.

\subsection{Definition of the Quality Metric $q_a$}

Finally, we tested different ways to define $q_a$. We considered two possibilities: 
\begin{equation*}
    q_a(D) = max(q_{a,1}(D), q_{a,2}(D))
\end{equation*} and 
\begin{equation*}
    q_a(D) = \alpha q_{a,1}(D) + (1-\alpha)q_{a,2}(D) (\text{with }0 \leq \alpha \leq 1)
\end{equation*}

By definition, $q_{a,1}$ and $q_{a,2}$ are bounded between 0 and 1. They indicate good quality when they are close to 0 and bad quality when they are close to 1. We expect $q_a$ to keep these same properties. Besides, $q_a$ should increase when the number of missing values or outliers injected in data increases and stay quite steady when fuzzing is injected. 

\begin{figure*}[!h]
    \hspace{-0.5cm}
    \centering
    \begin{subfigure}{0.35\textwidth}
        \includegraphics[width=\textwidth]{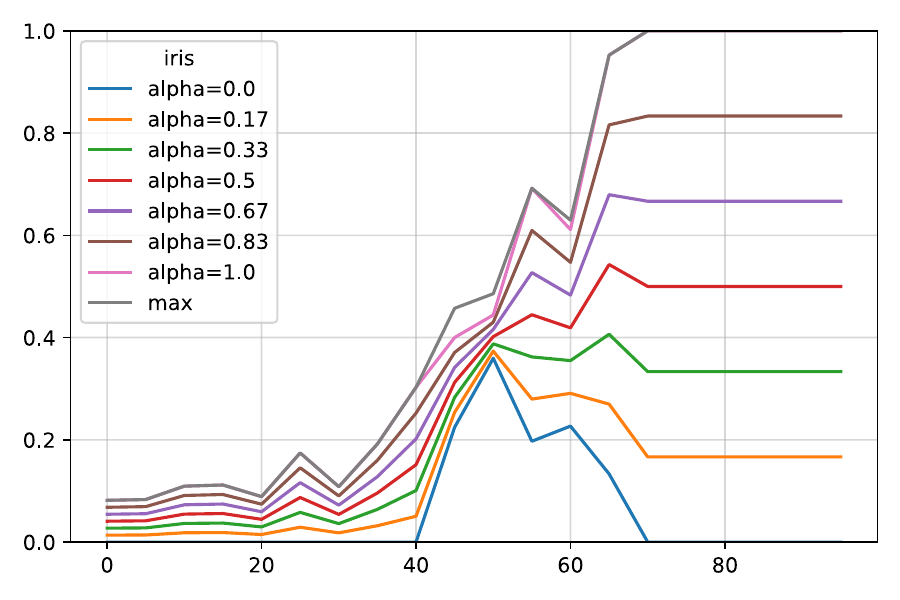}
        \vspace*{-0.6cm}
        \caption{}
        \label{ymiris}
    \end{subfigure}
    \hspace{-0.35cm}
    \begin{subfigure}{0.35\textwidth}
        \includegraphics[width=\textwidth]{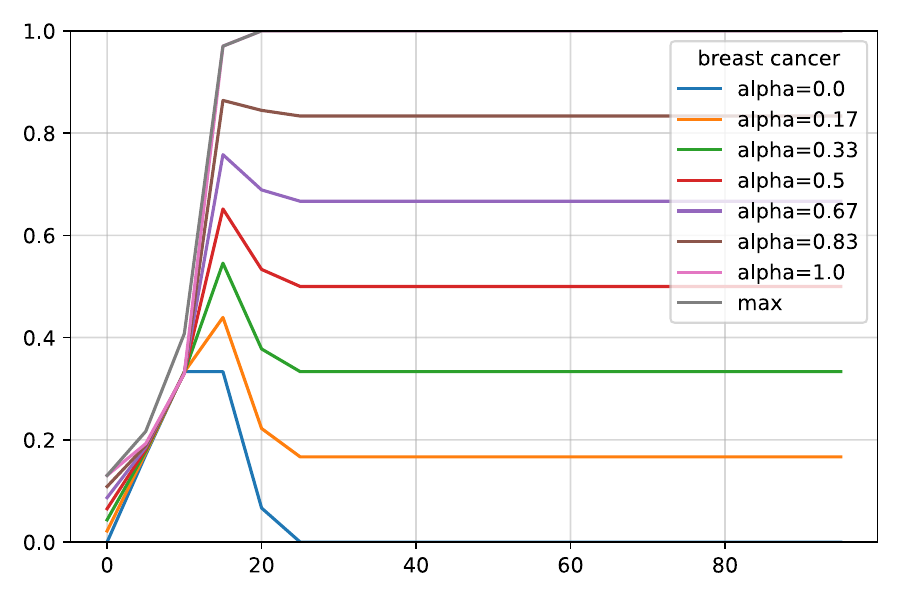}
        \vspace*{-0.6cm}
        \caption{}
        \label{ymcancer}
    \end{subfigure}
    \hspace{-0.35cm}
    \begin{subfigure}{0.35\textwidth}
        \includegraphics[width=\textwidth]{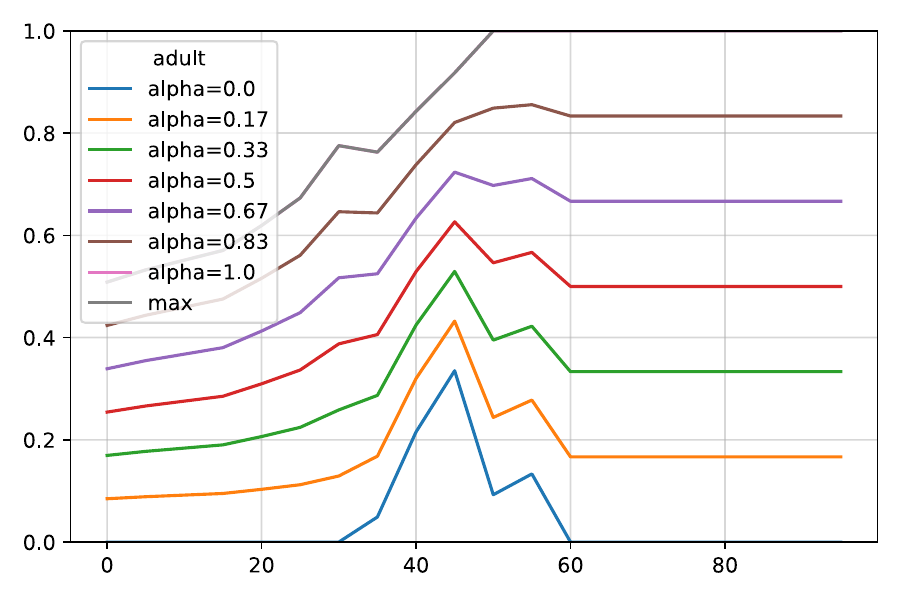}
        \vspace*{-0.6cm}
        \caption{}
        \label{ymadult}
    \end{subfigure}
    \hspace{-0.5cm}
    
    \hspace{-0.5cm}
    \centering
    \begin{subfigure}{0.35\textwidth}
        \includegraphics[width=\textwidth]{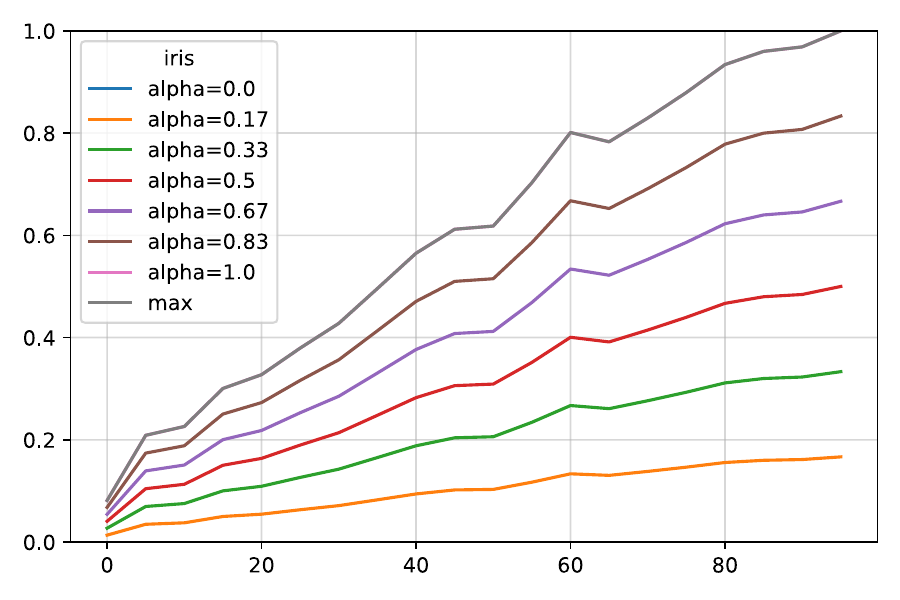}
        \vspace*{-0.6cm}
        \caption{}
        \label{yoiris}
    \end{subfigure}
    \hspace{-0.35cm}
    \begin{subfigure}{0.35\textwidth}
        \includegraphics[width=\textwidth]{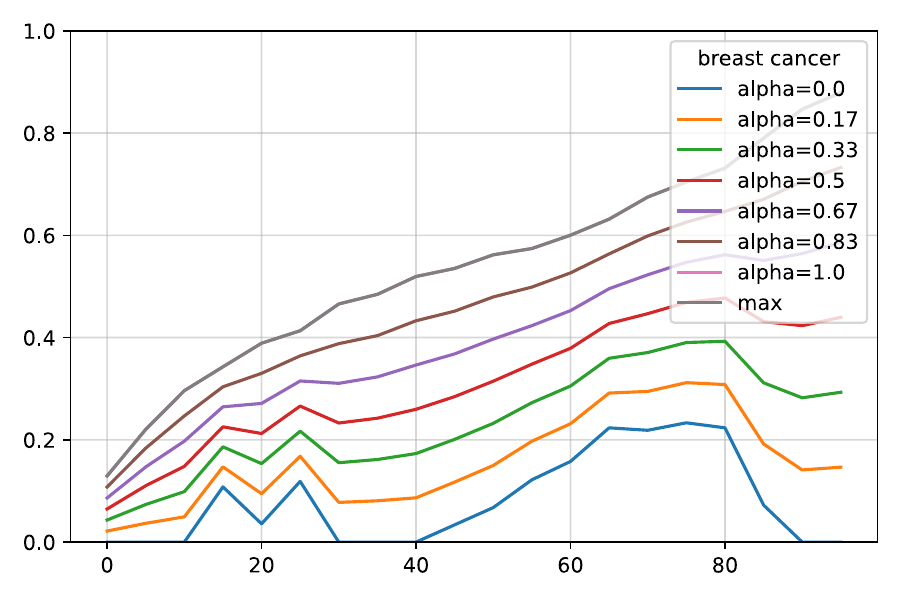}
        \vspace*{-0.6cm}
        \caption{}
        \label{yocancer}
    \end{subfigure}
    \hspace{-0.35cm}
    \begin{subfigure}{0.35\textwidth}
        \includegraphics[width=\textwidth]{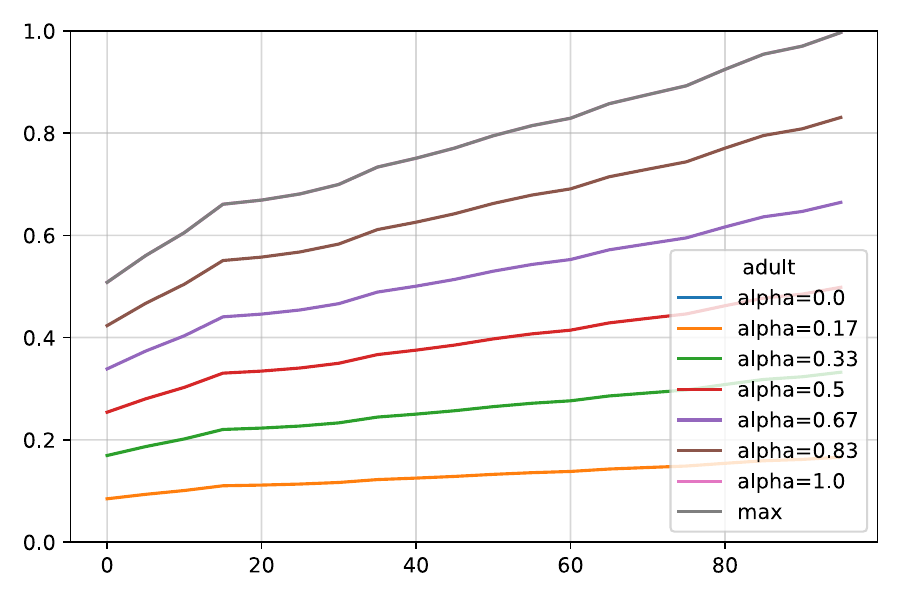}
        \vspace*{-0.6cm}
        \caption{}
        \label{yoadult}
    \end{subfigure}
    \hspace{-0.5cm}
    
    \hspace{-0.5cm}
    \centering
    \begin{subfigure}{0.35\textwidth}
        \includegraphics[width=\textwidth]{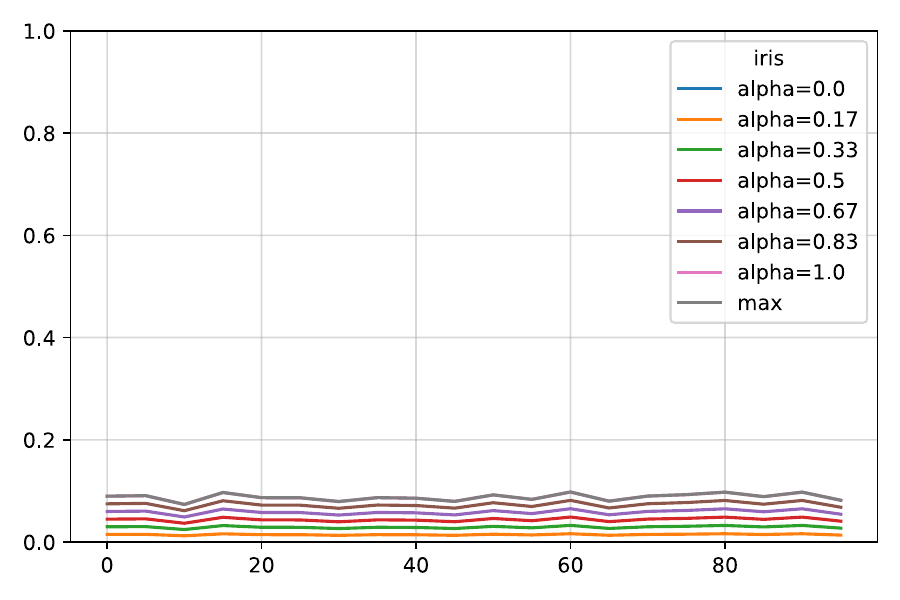}
        \vspace*{-0.6cm}
        \caption{}
        \label{yfiris}
    \end{subfigure}
    \hspace{-0.35cm}
    \begin{subfigure}{0.35\textwidth}
        \includegraphics[width=\textwidth]{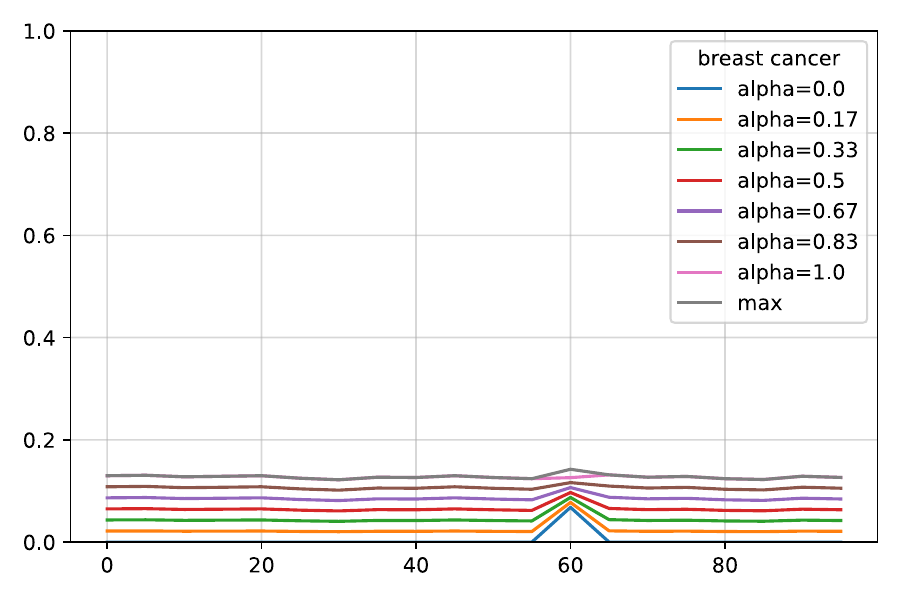}
        \vspace*{-0.6cm}
        \caption{}
        \label{yfcancer}
    \end{subfigure}
    \hspace{-0.35cm}
    \begin{subfigure}{0.35\textwidth}
        \includegraphics[width=\textwidth]{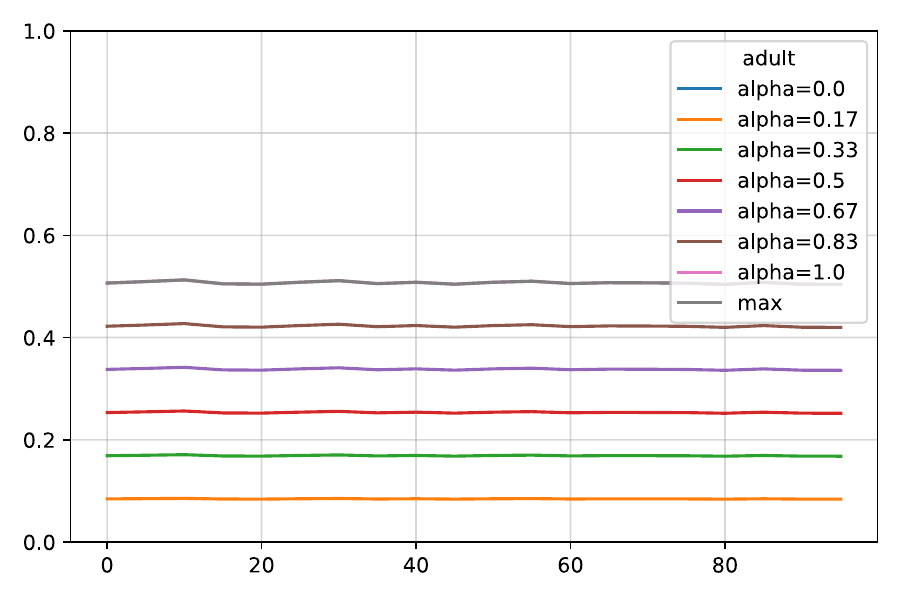}
        \vspace*{-0.6cm}
        \caption{}
        \label{yfadult}
    \end{subfigure}
    \hspace{-0.5cm}
\caption{Different computations of $q_a$ on datasets (columns): Iris (a, d, g), Breast Cancer (b, e, h), and Adult (c, f, i), when errors (lines): missing values (a, b, c), outliers (d, e, f), and fuzzing (g, h, i) are injected in data.}
\label{ycalib}
\vspace{-0.2cm}
\end{figure*}

To study the most relevant definition, we computed $q_a$ again for our three reference datasets and their deteriorated versions when up to 95\% of errors were injected in 5\% increments. Figure \ref{ycalib} presents $q_a$ computed with these possible definitions. As discussed in Section \ref{badtestdata}, $q_a$ is computed with the means of $q_{a,1}$ and $q_{a,2}$ over 30 resamplings to account for possible bad choices of testing data. 

In Figure \ref{ycalib}, we observe $q_a$ for Iris, Breast Cancer, and Adult, respectively, in columns 1, 2, and 3 as missing values, outliers, and fuzzing are injected respectively in rows 1, 2, and 3. For instance, in Figure \ref{ymiris}, we observe different ways to compute $q_a$ on the datasets obtained from Iris with incremental injection of missing values. 

We observe decreases in $q_a$ for the definitions $q_a = \alpha q_{a,1} + (1-\alpha)q_{a,2}$ in Figure \ref{ymiris}, \ref{ymcancer}, \ref{ymadult}, and \ref{yocancer} while the percentage of errors injected in data increases. This is a result of $q_{a,1}$ being constantly high since the mean accuracy is close to or lower than $\frac{1}{c}$, and $q_{a,2}$ being close or equal to 0 as the variations of accuracy are extremely low. This is problematic since we expect $q_a$ to increase until it reaches 1. This unexpected observation is no more visible when $q_a$ is defined as $q_a=max(q_{a,1}, q_{a,2})$. Additionally, we believe the maximum will capture more quality variations and thus be more sensitive. We define $q_a$ as:
\abovedisplayskip=3pt
\belowdisplayskip=3pt
\begin{equation}
	q_a(D) := max(q_{a,1}(D), q_{a,2}(D))
	\label{defqa}
\end{equation}

 We are now ready to present, in Algorithm \ref{algoqa}, all the steps required for evaluating the quality of $D$ with $q_a(D)$. If no trusted test is available, we compute $q_a(D)$ over 30 resamplings of test and train as studied in Section \ref{badtestdata}. $q_{a,1}$ and $q_{a,2}$ are then computed as the means of the $q_{a,1}$ and $q_{a,2}$ obtained from the resampled datasets.
\begin{algorithm}[!h]
	\hrulefill
	\caption{$q_a(D)$ computation}
	\label{algoqa}
	\KwData{Dataset $D$}
	\KwResult{$q_a(D)$ and its interpretation.}
	\eIf{$D$ is made up of a trusted test dataset}{
		$LD = ( D )$\;
	}
	{
		Generate the list $LD = ( D^1, \dots, D^{30} )$ of resampled versions of $D$\;
	}
	
	\ForEach{$D_i\in LD$}{

			Compute $q_{a,1}(D^i)$ as defined in Eq. \ref{basepartacc}\;
			\ForEach{ error type $e \in E$}{
				Create a new dataset $D^i_{e,p}$ by injecting $D^i$ with $p\%$ of error $e$, randomly  generated with a uniform distribution\;
			}
			Compute $q_{a,2}(D^i)$ with Eq.(\ref{varpartacc})\;
		}
		
		$q_{a,1}(D) = \frac{1}{30}\sum\limits_{i=1}^{30} q_{a,1}(D^i)$\;
		$q_{a,2}(D) = \frac{1}{30}\sum\limits_{i=1}^{30} q_{a,2}(D^i)$\;
		Compute $q_a(D)$ with Eq.(\ref{defqa})\;
		Interpret $q_a(D)$\;
	
	\hrulefill
\end{algorithm}

\subsection{Interpretation}
\label{zonesqa}
In this section, we discuss the interpretation of $q_a$. More specifically, we propose to extract thresholds for $q_a$ that express the notions of good, medium, or bad quality. To do this, we computed $q_a$ on 114 datasets derived from Iris, Breast Cancer, and Adult when errors are injected in 5\% increments from 0\% to 95\%. We only considered the error types missing values and outliers, as fuzzing does not tend to affect classifier performance. Since we control the level of errors, we estimate the different levels of quality for these datasets. Iris and Breast Cancer are considered good quality when used with classification models in the literature, while the Adult is of medium quality. We expect data quality to be good for the datasets obtained from  Iris and Breast Cancer with up to 10\% of errors, medium between 10\% and 30\%, and bad for over 30\% of errors. For the datasets obtained from Adult, we expect medium quality up to 10 \% of errors and then bad quality.
 
\begin{figure*}[!h]
    \begin{center}
    \begin{subfigure}{0.35\textwidth}
        \includegraphics[width=\textwidth]{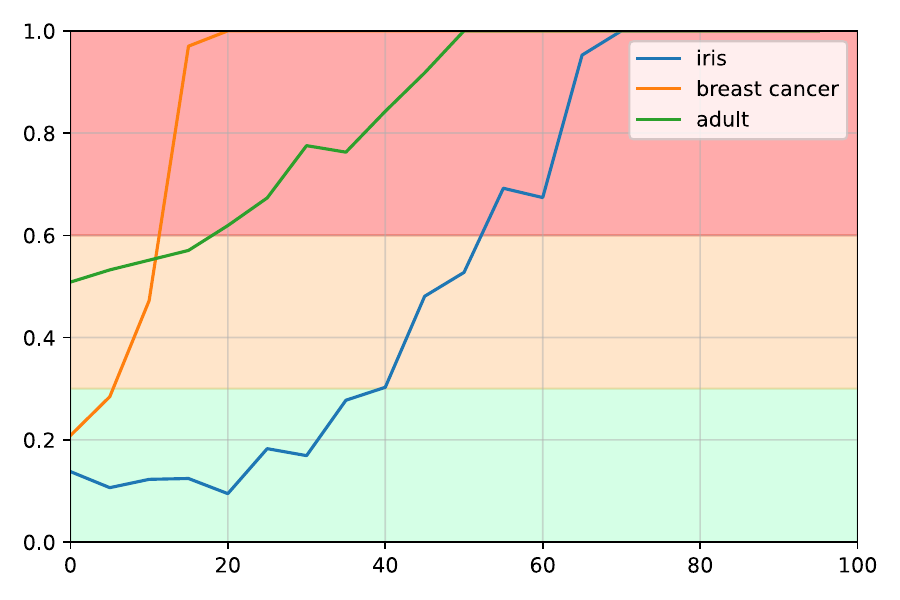}
        \vspace*{-0.6cm}
        \caption{}
        \label{qam}
    \end{subfigure}
    \begin{subfigure}{0.35\textwidth}
        \includegraphics[width=\textwidth]{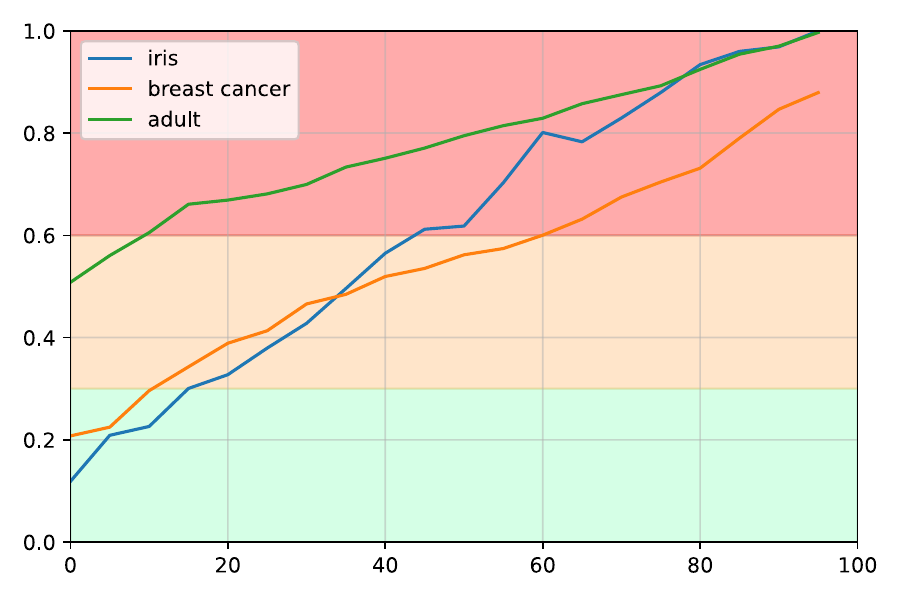}
        \vspace*{-0.6cm}
        \caption{}
        \label{qao}
    \end{subfigure}
    \end{center}
    \caption{$q_a$ when missing values (a) and outliers (b) are injected in the datasets.}
    \label{ydet}
    \vspace{-0.2cm}
\end{figure*}
We present, in the figures \ref{qam}, \ref{qao}, the evolution of $q_a$ for three dataset examples when controlled percentages of missing values, outliers, and fuzzing are respectively injected with 5\% increments. Colors on the $y$ axis depict the proposed thresholds for which $q_a$ indicates good, medium, or bad data quality.

These thresholds were empirically chosen based on the following observations. In Figure \ref{ydet}, we see that without any deterioration, $q_a$ measurements for these datasets are respectively 0.11, 0.2, and 0.5. Thus, we expect  0.11 and 0.2 to indicate good quality and 0.5 to indicate medium quality. Furthermore, in Figure \ref{qao} for around 10\% of errors, when data quality is expected to be altered from good to medium, we observe that values of $q_a$ are close to 0.3 for the datasets Breast Cancer and Iris. This prompts us to set the upper threshold for good quality at 0.3, which is consistent with our first observations. Additionally, in Figure \ref{accm}, we observe that the mean accuracy for the dataset Adult starts to decrease significantly between 15\% and 20\% of missing values injected. This corresponds to $q_a$ measurements between 0.55 and 0.7 in Figure \ref{accm}. From these elements, we propose to set the upper threshold for medium quality to 0.6. We find this to be a reasonable limit since, if we take the case of a dataset with two classes, being over this threshold either means that its mean accuracy is below 0.7 or that it experiences a mean variation of accuracy over 0.18 when injected with 5\% of errors. These requirements either indicate a mean accuracy low enough or variations of accuracy high enough to characterize a dataset as bad quality.

We, therefore, propose the following thresholds to interpret $q_a$: if $q_a(D) \leq 0.3$, this means that $D$ is of good quality; If $0.3 < q_a(D) \leq 0.6$ $D$ can be considered to be of medium quality, and examining the values of $q_{a,1}(D)$ and $q_{a,2}(D)$ is necessary to decide whether $D$ can be used. Finally, $0.6 < q_a(D)$ means that $D$ is of bad quality.

However, in Figure \ref{qam}, we can see that for 20\% of missing values, we measure $q_a(Iris)<0.3$. This result is unexpected since, for this percentage of errors, we do not expect good data quality. Nonetheless, it is consistent with Figure \ref{accm} where we can observe that at 20\% of missing values, accuracy is still high with Iris and is not about to experience a significant drop. 

\section{Empirical Evaluation}
\label{EmpEval}
The experiments presented in this section aim to evaluate the relevance of $q_a$. For the remainder of the paper, we express this relevance through 3 questions:
\begin{itemize}[noitemsep,topsep=0pt]
    \item Q1: Can $q_a$ characterize a dataset of good or medium quality?
    \item Q2: Can $q_a$ characterize a dataset of bad quality?
    \item Q3: Can $q_a$ provide valuable information to interpret a quality score?
\end{itemize}

\subsection{Empirical Setup}
\label{datasets}
First, we use the classification model set $M$ listed in Section \ref{qaintro} along with the three error types: missing values, outliers, and fuzzing. We also set the percentage of injected errors $p=5\%$ as discussed in Section \ref{qaintro}. 

We evaluated $q_a$ on 155 numeric datasets. We used five distinctive datasets: Spambase, Heart Disease, Abalone, Dry Beans, and Statlog \cite{uci}, as well as 150 datasets we modified by injecting controlled amounts of errors. We selected these datasets for their varied dimensions, number of classes, number of attributes, and domains of application. The objective of the Spambase dataset is to predict whether an email is a spam. The Heart Disease dataset is used to predict the presence of heart diseases in patients. The Abalone dataset is used for the prediction of the number of rings present in abalone shells (which indicates their ages) from physical measurements. The class imbalance in this dataset is too high to achieve reasonable accuracies. Therefore, we chose to work on a more straightforward classification task by aggregating the classes into two groups: up to 8 shell rings and over eight shell rings. The Dry Beans dataset is used for the prediction of the varieties of dry beans from features by the market situation. Finally, the dataset Statlog is used to classify people described by attributes as good or bad credit risks.

We manually evaluated the quality of these five datasets by measuring 7 dimensions and characteristics. The results, along with our quality estimations, are provided in Table \ref{tabdata}. The dataset Spambase is estimated to be of good quality as its number of samples (4 601) is relatively high even for its less populated class (1 813) compared to its number of attributes and classes (57 and 2). Besides, its mean accuracy across classification models is high (0.9). Datasets Abalone, Statlog, Dry Beans, and Heart Disease are estimated to be of medium quality mainly because their mean accuracies are lower (respectively 0.84, 0.76, 0.68, and 0.79). Additionally, the datasets Abalone and Dry Beans present high levels of class imbalances, which are usually considered quality issues. The mean accuracy for the dataset Dry Beans can seem very low compared to the other evaluation datasets, but in the context of a seven-class classification problem, it is, in fact, rather high. 

105 additional datasets were derived from the five previous ones by injecting missing values, outliers, and fuzzing, separately and randomly with a uniform distribution. Two strategies were followed: injection of 5 and 10\% of errors to build datasets of good or medium qualities, and injection of 30 up to 50\% of errors with increments by 5\% to build datasets of bad quality. Datasets, results, and a prototype version of our tool allowing to compute $q_a$ are available in \cite{CompanionSite}.

\begin{table*}
\caption{Overview of the evaluation datasets}
\begin{adjustbox}{width=\textwidth, center}
        \begin{tabular}{|c|c|c|c|c|c|c|c|c|}\hline
            \textbf{Dataset} & \textbf{\makecell{Number of \\ classes}} & \textbf{\makecell{Samples total}} & \textbf{\makecell{Number of \\ attributes}} & \textbf{Features} & \textbf{Class imbalance} & \textbf{\makecell{Missing \\ data}} & \textbf{\makecell{Mean \\ accuracy}} & \textbf{\makecell{Estimated \\ data quality}}\\ \hline
            Spambase & 2 & 4 601 & 57 & integers, reals & \makecell{Yes (1 813 samples \\ in the least populated class)} & None & 0.90 & good\\ \hline
            Abalone & \makecell{28 \\ (2 post-processing)} & 4 177 & 8 & \makecell{categorical,\\ integers, reals} & \makecell{Yes (1 407 samples \\ in the least populated class \\ post-processing)} & None & 0.84 & medium\\ \hline
            Dry Beans & 7 & 13 611 & 16 & \makecell{categorical,\\ integers, reals} & \makecell{Yes (522 samples \\ in the least populated class)} & None & 0.68 & medium\\ \hline
            Statlog & 2 & \makecell{1 000 \\ (959 post-processing)} & 23 & integers & \makecell{Yes (275 samples \\ in the least populated class)} & \makecell{on 41 \\ samples} & 0.76 & medium\\ \hline
            Heart Disease & \makecell{5 \\ (2 post-processing)} & \makecell{303 \\ (297 post-processing)} & 13 & \makecell{categorical,\\ integers, reals} & \makecell{Yes (No after \\ post-processing)} & \makecell{on 6 \\ samples} & 0.79 & medium\\ \hline
        \end{tabular}
\end{adjustbox}
    \label{tabdata}
    \vspace{-0.2cm}   
\end{table*}

\subsection{Q1: Can $q_a$ Characterize a Dataset of Good or Medium Quality?}
\label{q1}
To answer this question, we computed $q_a$ for the five datasets presented in Table \ref{tabdata} as well as for the 30 datasets created by injecting 5\% and 10\% of missing values, outliers, or fuzzing.   
In Table \ref{tabqa}, we present $q_a$ for the five datasets without deterioration, $q_a$ is presented with the details of $q_{a,1}$,  $q_{a,2}$, and the corresponding data quality levels defined in Section \ref{zonesqa}. $q_a$ was computed over the 12 classification models and 30 resamplings of training and test data. We compare these quality levels with the data quality estimations given in Table \ref{tabdata}.

\begin{table}
\caption{$q_a$ calculation and the related quality levels}
\begin{adjustbox}{width=0.8\columnwidth, center}
    \begin{tabular}{|c|c|c|c|c|}\hline
        \textbf{Dataset} & $\mathbf{q_a}$ & $\mathbf{q_{a,1}}$ & $\mathbf{q_{a,2}}$ & \textbf{Quality level}\\ \hline
         Spambase & 0.29 & 0.18 & 0.29 & good\\ \hline
         Abalone & 0.32 & 0.32 & 0.02 & medium\\ \hline
         Dry Beans & 0.36 & 0.36 & 0.17 & medium\\ \hline
        Statlog & 0.48 & 0.48 & 0.13 & medium\\ \hline 
        Heart Disease & 0.42 & 0.42 & 0.11 & medium\\ \hline

    \end{tabular}
\end{adjustbox}
    \label{tabqa}
    \vspace{-0.2cm}   
\end{table}

We observe with Tables \ref{tabqa} and \ref{tabdata} that $q_a$ evaluates the data quality of the five first datasets correctly. The only dataset evaluated as of good quality by $q_a$ is Spambase. We can also see that its values for $q_{a,1}$ and $q_{a,2}$ are low. This means that the classification models perform with high levels of accuracy ($q_{a,1}$) and that these performances would not drop significantly upon small perturbations of the dataset ($q_{a,2}$). Indeed, its mean accuracy is high (0.9), and even if the dataset presents a class imbalance, its least populated class is still relatively populated (1 813 samples). The other datasets in Table \ref{tabqa} are all classified as medium quality by $q_a$. Again, this is consistent with our estimated data qualities given in Table \ref{tabdata}. The quality issues are different, though. For instance, for the dataset Abalone, $q_{a,1}$ is relatively high, but $q_{a,2}$ is low. This means that the accuracy across models is not very high, but its accuracy does not vary much when 5\% of errors are injected in the training set. Moreover, the mean accuracy across classification models for the dataset Abalone is 0.84 (Table \ref{tabdata}). Since the dataset only has two classes, this is not considered a very high accuracy. However, depending on the application, it can be regarded as reasonable. %For the dataset Dry Beans, $q_a$ expresses a similar level of quality with $q_a(dry\ beans) = 0.36$, and $q_a(Abalone) = 0.32$. The datasets Heart Disease and Statlog are also evaluated by $q_a$ as of medium quality with $q_a(heart\ disease) = 0.42$, and $q_a(Statlog)=0.48$. This is consistent with the data quality given in Table \ref{tabdata}.  

\begin{figure*}[!h]
    \hspace{-0.5cm}
    \centering
    \begin{subfigure}{0.35\textwidth}
        \centering
        \includegraphics[width=\textwidth]{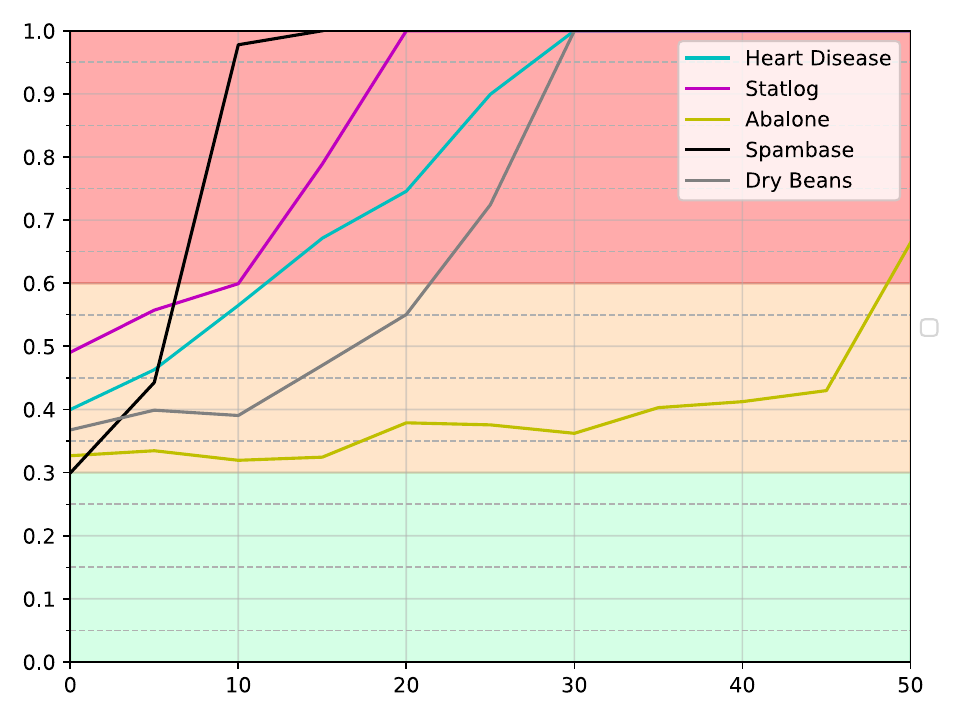}
        \vspace*{-0.6cm}
        \caption{}
        \label{qadm}
    \end{subfigure}
    \hspace{-0.35cm}
    \begin{subfigure}{0.35\textwidth}
        \centering
        \includegraphics[width=\textwidth]{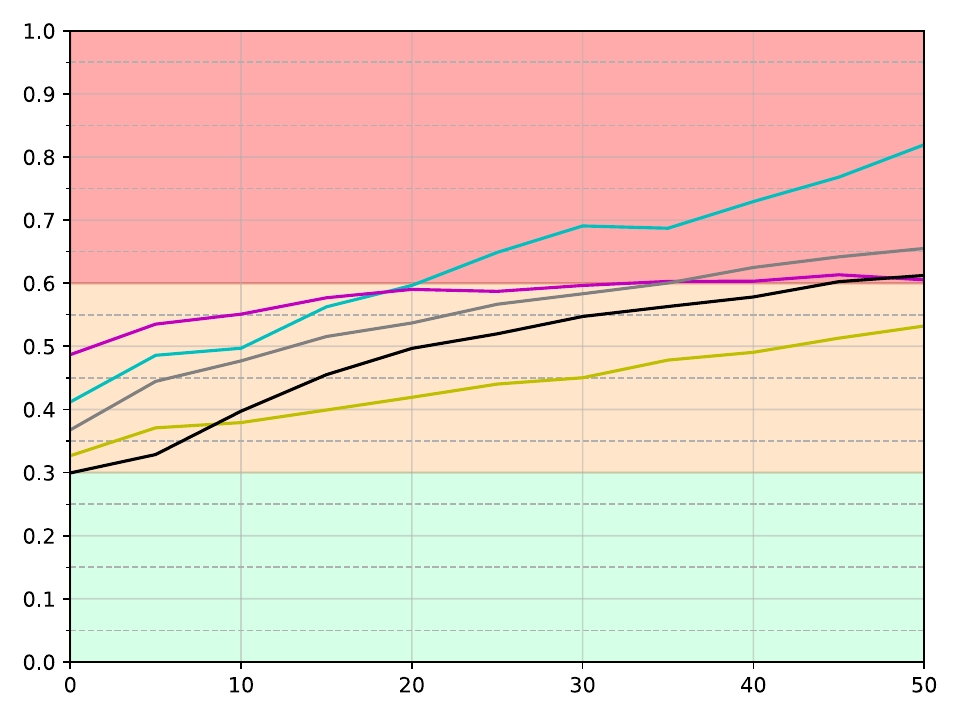}
        \vspace*{-0.6cm}
        \caption{}
        \label{qado}
    \end{subfigure}
    \hspace{-0.35cm}
    \begin{subfigure}{0.35\textwidth}
        \centering
        \includegraphics[width=\textwidth]{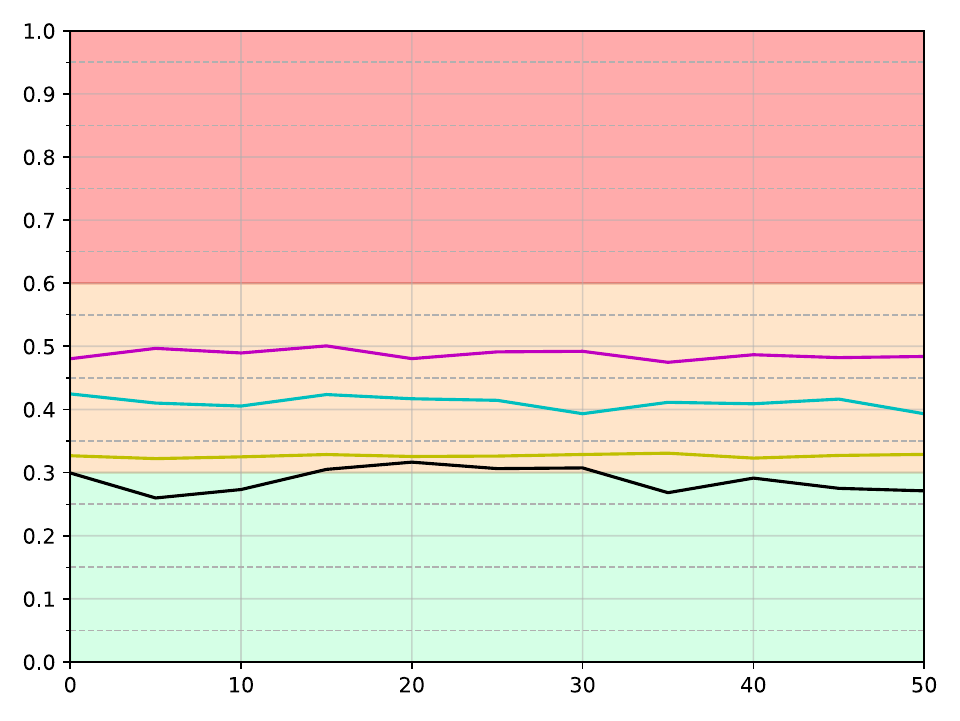}
        \vspace*{-0.6cm}
        \caption{}
        \label{qadf}
    \end{subfigure}
    \hspace{-0.5cm}
    \caption{$q_a$ when missing values (a), outliers (b), and fuzzing (c) are injected in all datasets.}
    \label{qaeval}
    \vspace{-0.2cm}   
\end{figure*}

Figure \ref{qadm}, \ref{qado}, and \ref{qadf} now respectively illustrate the values of $q_a$ when injecting missing values, outliers, and fuzzing in the previous datasets with a 5\% increment from 0 to 50\%. For this question, we only focus on the 30 datasets obtained when 5\% and 10\% of errors are injected. We expect them to be of medium data quality. The colored zones in Figure \ref{qaeval} correspond to the thresholds we set in Section \ref{zonesqa}, for which quality is considered good, medium, or bad.

We observe in these figures that $q_a$ indicates a good or medium quality level for 28 of these 30 datasets. In Figure \ref{qadm}, $q_a$ indicates a bad quality for the datasets Spambase and Statlog at 10\% of missing values injected. For the dataset Statlog, with 10\% of missing values, we observed that the mean accuracy score is equal to 0.7. This dataset also presents a class imbalance that can be worsened by the injection of missing values. Its quality evaluation is, therefore, consistent with our observations. For Spambase, the mean accuracies of the datasets with 0, 5, and 10\% of missing values are 0.9, 0.85, and 0.56. These results tend to show that the dataset with 10\% of errors is of bad quality for classification tasks. In Figure \ref{qado}, we observe that $q_a$ indicates medium data quality for all datasets between 0\% and 10\% of missing values, which is consistent with our expectations. 

It is worth noting that, in Figure \ref{qadf}, $q_a$ stays relatively constant, which is what we expect.

Out of the 35 datasets studied in this section, $q_a$ is indicative of either good or medium data quality for 33 of them, which is the result we expected. For the 2 remaining cases where $q_a$ scores are indicative of bad quality, we studied carefully the datasets and observed that $q_a$ is correct. Thus, we conclude that $q_a$ is able to characterize datasets of good or medium quality. 

\subsection{Q2: Can $q_a$ Characterize a Dataset of Bad Quality?}
To investigate this question, we computed $q_a$ for the 50 datasets obtained after injecting controlled percentages of missing values and outliers in the initial datasets presented in Table \ref{tabdata}. We injected from 30 to 50\% of errors with a 5\% increment. At this level of errors for missing values and outliers, we assumed that the 50 resulting datasets were of bad quality. We do not consider fuzzing here as we expected that the injection of this error type should keep the quality close to the quality of the original datasets. The computed $q_a$ scores are again presented in Figure \ref{qaeval}.

Out of the 50 datasets studied in this section, $q_a$ indicated bad data quality for 37 of them. For example, we can see in Figure \ref{qadm} that when 30\% of missing values are present in datasets, $q_a$ correctly indicates bad data quality, except for the dataset Abalone up to 45\%. We observe similar results for outliers. We hence investigated why $q_a$ indicates medium data quality for the 9 datasets obtained from Abalone after injecting 30 to 45\% of missing values or outliers. The mean accuracies measured from these datasets stay over 75\% and then drop (abruptly in Figure \ref{qadm}). This is actually consistent with the Abalone properties given in Table \ref{tabdata}, where we observe that Abalone has a high number of samples (4 177) for a low number of classes and attributes (2 and 8). Hence, our own interpretation of quality was wrong for these 9 datasets. We can indeed estimate that these datasets are of medium quality with regard to these characteristics. The mean accuracy of the dataset Spambase stays over 0.7 up until the injection of 45\% of outliers. Its quality evaluation without the injection of errors was good (as opposed to medium for the other evaluation datasets). This could explain its higher resistance to outliers. Moreover, for the dataset Dry Beans at 30\% of outliers, the mean accuracy is evaluated at 0.5, which seems low but is still much better than a random guess with seven classes ($\frac{1}{7}$). 

In summary, $q_a$ indicates bad data quality for 37 datasets. For the 13 cases where $q_a$ indicates medium data quality when bad data quality was expected, we observed that the data characteristics and mean accuracies were consistent with a medium data quality level. These observations allow us to conclude that $q_a$ is able to characterize datasets of bad quality.

\subsection{Q3: Can $q_a$ Provide Valuable Information to Explain Data Quality?}
$q_a$ provides information to explain data quality in 3 ways.

Firstly, $q_a$ is computed across different classification models. This means that the resulting measures are not model-dependent. For instance, if we evaluate $q_a$ using only Ada-Boost, we have $q_a=0.34$, which corresponds to medium quality. If $q_a$ is evaluated with KNN, $q_a=0.62$, which corresponds to bad quality. However, as $q_a$ is evaluated across the 12 classification models presented in Section \ref{introduction}, $q_a=0.42$ (medium quality), as shown in Table \ref{tabqa}. $q_a$ provides general quality information. 

Secondly, $q_a$ brings valuable information, allowing us to compare accuracies for classification problems with datasets having different numbers of classes. For instance, Dry Beans has seven classes, while datasets Abalone and Spambase have two classes (Table \ref{tabdata}). The mean accuracy across classification models for the dataset Dry Beans is 0.68, while it is 0.84 for Abalone and 0.90 for Spambase. We can compare accuracies for Abalone and Spambase, but if we try to compare them with Dry Beans, we would conclude that the dataset Dry Beans is of lesser quality than Abalone and Spambase. However, the two classification problems are very different. Indeed, the accuracy corresponding to a random guess for the dataset Dry Beans is $\frac{1}{7}$, while it is $\frac{1}{2}$ for the datasets Abalone and Spambase. $q_{a,1}$ captures this comparison with random guess and helps compute quality measures, which are not dependent on the class number. In our case, we observe closer values of $q_{a,1}$ for Dry Beans and Abalone, while Spambase stays better than Abalone ($q_{a,1}(DryBeans)=0.36$, $q_{a,1}(Abalone)=0.32$, $q_{a,1}(Spambase)=0.18$). 

Finally, $q_a$ takes into account variations of accuracy with $q_{a,2}$. This allows us also to study model performance evolution when small amounts of errors are injected into a dataset. In other terms, this helps study whether models are resilient to errors. A model $m$ is said to be resilient when its performance stays consistent even if the dataset $D$ used for training $m$ is injected with a small amount of errors. This viewpoint is particularly interesting for online (a.k.a. continuous) machine learning, where models are continuously fed with new data that complete a dataset. For instance, if we compare the datasets Abalone and Spambase only with the information mentioned so far, we have respectively $q_{a,1}=0.32$ for Abalone and $q_{a,1}=0.18$ for Spambase. However, as shown by its $q_{a,2}$ value (Table \ref{tabqa}), Abalone is very resilient to  errors ($q_{a,2}=0.02$). This is not true for Spambase since $q_{a,2}=0.29$. Comparing these two datasets on accuracies alone would lead us to conclude that Spambase is of significantly better quality than Abalone. However, this only considers a situation where data is used at a specific point in time and never evolves. $q_{a,2}$, however, contains information that is valuable to predict a possible evolution of data quality if the data slightly deteriorates. 

This shows that $q_a$ does provide valuable information to explain data quality. Firstly, across classification models; secondly, for classification problems with datasets having different numbers of classes; and finally, on the resilience to errors of models. However, key information is still missing. For instance, the model performance is only evaluated with accuracy. In some situations, e.g., when datasets have imbalanced classes, accuracy is not the best measure. Furthermore, although $q_a$ seems to be a good way to measure a global quality level, it is not sufficient to explain this level. Other information, such as the completeness or class balance, would also be helpful in explaining data quality scores. For instance, a high level of class imbalance, as we see for the dataset Dry Beans in Table \ref{tabdata}, is not reflected by $q_a$.

\subsection{Threats to validity}
\label{threats}

In this section, we address 8 possible threats to the validity of this study. We identified 4 internal threats and 4 external threats. 

The internal threats we identified are the implementation of the classification models, the hyper-parametrization of the classification models, the number of datasets used to study the impact of untrusted test data, and the number of datasets used to define the thresholds to interpret $q_a$. To address the first threat, we implemented classification models using scikit-learn \cite{scikit-learn}, a widely used library. To limit the second threat, we used a grid search to set the hyper-parameters for classification models on all datasets without any deterioration and then used these settings for the rest of the experiments. To limit the third and fourth threats, we selected three widely used datasets, along with the 153 datasets obtained by injecting controlled percentages of errors in these 3 datasets.

The 4 external threats we identified are the choice of the datasets for this evaluation, the choice of classification models, the generation of errors, and the combination of errors. We tried to limit the first threat by choosing datasets that are widely used and have different dimensions; we also selected datasets that cover various applications and ranges of dimensions. We selected a wide range of classification approaches to address the second threat. To limit the third one, we decided to generate errors randomly, with a uniform distribution in datasets. Finally, we decided to study errors separately to limit the fourth threat. However, we plan to extend this work to error combinations in future work.

\section{Conclusion}
\label{conclusion}
In this paper, we have introduced a novel metric to measure data quality. The main advantage of the proposed metric is being independent of learning models and expert knowledge. Furthermore, it does not require external reference data. As a consequence, it offers the possibility to compare different datasets. We have extensively tested and evaluated the proposed metric and have shown that it is able to characterize data quality correctly.

\bibliographystyle{apalike}
{\small
\bibliography{biblio}}

\end{document}